\documentclass[11pt,a4paper]{article}

\usepackage[a4paper,top=2.25cm,bottom=2.35cm,left=2.35cm,right=2.35cm]{geometry}
\usepackage[T1]{fontenc}
\usepackage{lmodern}
\usepackage{microtype}
\usepackage{amsmath,amssymb,amsfonts,bm}
\usepackage{booktabs,multirow,array,tabularx}
\usepackage{graphicx}
\usepackage{xcolor}
\usepackage{tikz}
\usepackage{pgfplots}
\usepackage{algorithm}
\usepackage{algpseudocode}
\usepackage{enumitem}
\usepackage{caption}
\usepackage{subcaption}
\usepackage{fvextra}
\usepackage{siunitx}
\usepackage{authblk}
\usepackage{fancyhdr}
\usepackage{titlesec}
\usepackage{xurl}
\usepackage{hyperref}
\usepackage[nameinlink,noabbrev]{cleveref}
\usepackage[backend=biber,style=ieee,sorting=none]{biblatex}

\usetikzlibrary{positioning,arrows.meta,calc}
\pgfplotsset{compat=1.18}
\definecolor{paperblue}{HTML}{1F4E79}
\definecolor{accent}{HTML}{0F9D9A}
\definecolor{warmgold}{HTML}{C9942B}
\definecolor{textgray}{HTML}{3B4652}
\definecolor{lightbg}{HTML}{F7F9FB}

\hypersetup{
  colorlinks=true,
  linkcolor=paperblue,
  citecolor=paperblue,
  urlcolor=paperblue,
  pdftitle={Reinforcement Fine-Tuning for Signal Mathematical Reasoning},
  pdfauthor={Guozheng Sun}
}

\titleformat{\section}{\large\bfseries\color{paperblue}}{\thesection}{0.7em}{}
\titleformat{\subsection}{\normalsize\bfseries\color{textgray}}{\thesubsection}{0.7em}{}
\titleformat{\subsubsection}{\normalsize\itshape\color{textgray}}{\thesubsubsection}{0.7em}{}
\titlespacing*{\section}{0pt}{1.6ex plus 0.4ex minus 0.2ex}{0.8ex}
\titlespacing*{\subsection}{0pt}{1.2ex plus 0.3ex minus 0.2ex}{0.55ex}

\RecustomVerbatimEnvironment{verbatim}{Verbatim}{breaklines=true,breakanywhere=true,fontsize=\small}
\setlist[itemize]{topsep=3pt,itemsep=2pt,parsep=0pt,leftmargin=1.3em}
\setlist[enumerate]{topsep=3pt,itemsep=2pt,parsep=0pt,leftmargin=1.5em}

\newcommand{\model}{Qwen2.5-3B-Base}
\newcommand{\domain}{signal processing}
\newcommand{\bench}{WirelessMATHBench-XL}
\newcommand{\sftdata}{Wireless-CoT-Mix}

\newcommand{\keywords}{Signal Mathematical Reasoning, SFT, GRPO, GSPO, GMPO}

\title{\vspace{-1.5cm}
{\Large\bfseries SignalReasoner: Assessing the Upper Bound of 3B Models for Signal Mathematical Reasoning}\\[0.45em]
}
\author[]{Guozheng Sun}
\date{}

\begin{document}
\maketitle
\thispagestyle{plain}
\vspace{-0.8cm}
\begin{abstract}
\noindent
Post-training with supervised chain-of-thought fine-tuning and reinforcement learning from verifiable rewards has substantially improved the mathematical reasoning capabilities of large language models (LLMs).
However, their application to \domain{} problems remains relatively under-explored.
This report investigates reinforcement fine-tuning strategies for adapting \model{} to graduate-level signal mathematical problems from \bench{}, a comprehensive benchmark for mathematical reasoning in this domain.
We examine two training paradigms: (i) direct reinforcement learning (RL) on \bench{} with verifiable rewards; and (ii) supervised fine-tuning (SFT) on a distilled wireless-domain chain-of-thought corpus, followed by the same domain-specific RL stage. Across both paradigms, we benchmark Group Relative Policy Optimization (GRPO), Group Sequence Policy Optimization (GSPO), and Geometric-Mean Policy Optimization (GMPO). 
We aim to assess whether domain-aware CoT SFT serves as an effective initialization for subsequent RL, and whether GSPO or GMPO offer advantages in stability or accuracy over GRPO for signal reasoning tasks. 
Our best model achieves an overall accuracy of 39.12\%, representing a more than threefold improvement over the untrained Base model (12.37\%).
\end{abstract}

\vspace{0.4em}

\vspace{0.35em}
\noindent\textbf{Code:} \url{https://github.com/thusunlight/SignalReasoner}

\begin{figure}[H]
  \includegraphics[width=\textwidth]{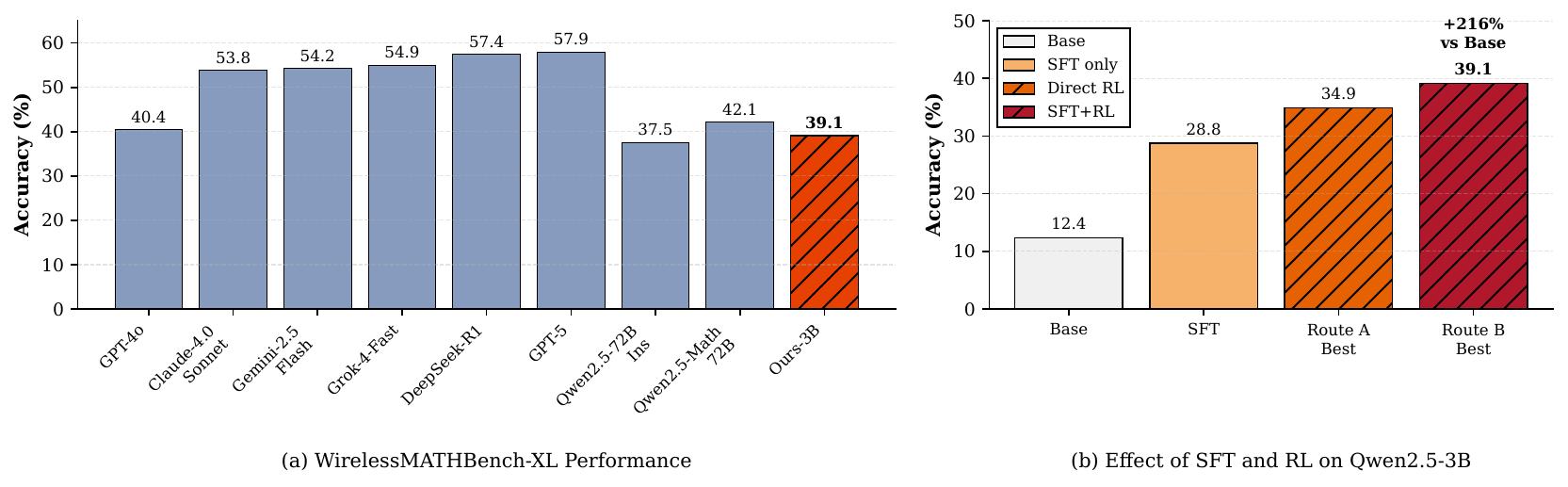}
  \caption{\small Teaser comparison on \bench{}. Blue bars show reference strong-model results, while orange/red bars summarize our Qwen2.5-3B experiments. Route~A denotes direct RL, and Route~B denotes SFT+RL.}
\end{figure}
\vspace{-0.4em}

\section{Introduction}
Large language models (LLMs) have shown strong progress in reasoning~\cite{wei2022chain,kojima2022zero,wang2023self,gong2026replanbot}, especially when post-trained with supervised chain-of-thought data and reinforcement learning from verifiable rewards~\cite{zelikman2022star,shao2024deepseekmath,deepseek2025r1}. However, specialized engineering problems remain difficult for general-purpose models. 
Signal processing questions involve technologies such as MIMO and beamforming~\cite{molisch2022wireless}, RIS~\cite{basar2019ris}, and ISAC~\cite{liu2022isac}, and often combine symbolic equations, physical assumptions, and long scientific context. 
A correct answer may require not only algebraic manipulation, but also recognition of channel models, signal dimensions, 
and the meaning of variables inside a system equation. 
Existing sub-10B models struggle with such tasks, as they lack the capacity to jointly handle domain-specific knowledge and multi-step mathematical derivations~\cite{li2025wirelessmathbench,li2025wirelessmathlm,wei2022emergent}.

This report investigates whether Qwen2.5-3B~\cite{yang2024qwen25} can be adapted to this domain through reinforcement fine-tuning, and to what extent post-training improves its reasoning capability. We conduct experiments on \bench{}~\cite{li2025wirelessmathlm}, which consists of graduate-level signal mathematical problems derived from technical literature, and evaluate two training paradigms with three RL algorithms.
In the first route, \model{} is directly optimized with RL on \bench{}. This route tests whether the base model can acquire domain reasoning behavior from reward signals alone. In the second route, \model{} is first fine-tuned on \sftdata{}, a 3,542-example mixture of distilled wireless-domain CoT trajectories and NuminaMath-CoT~\cite{numina2024math} samples, and then further optimized with RL on \bench{}. This route tests whether domain-aware CoT SFT provides a better initialization for subsequent signal-domain RL.

For the RL stage, we compare three group-based policy optimization algorithms, namely GRPO~\cite{shao2024deepseekmath}, GSPO~\cite{zheng2025gspo}, and GMPO~\cite{zhao2025gmpo}. GRPO removes the need for a separate critic by estimating advantages from multiple responses sampled for the same prompt. GSPO changes the optimization unit from tokens to full sequences, aiming to improve training stability by aligning sequence-level rewards with sequence-level clipping. GMPO instead stabilizes token-level updates by replacing arithmetic aggregation with a geometric-mean objective that reduces sensitivity to outlier importance ratios.
Experiment results show that our best model achieves an overall accuracy of 39.12\%, representing a more than threefold improvement over the untrained Base model (12.37\%).

\paragraph{Contributions.}
This report makes the following contributions:
\begin{itemize}
  \item We conduct a systematic empirical comparison of two fine-tuning routes for adapting Qwen2.5-3B to \domain{} tasks on \bench{}, and demonstrate that RL consistently improves overall accuracy from 12\% to 39\%.
  \item We benchmark GRPO, GSPO, and GMPO across both routes, finding that GSPO and GMPO outperform GRPO in accuracy, convergence speed, and output token efficiency, with GMPO after SFT achieving the best overall result.
  \item We identify a reasoning-shortening phenomenon under GSPO and GMPO, where the model learns to produce compact answers with reduced chain-of-thought detail, and propose a plausible explanation for the underlying mechanisms of this behavior.
  \end{itemize}

\section{Related Work}

\subsection{Reinforcement Learning for Reasoning}
Aligning language models with human intent has driven the RLHF paradigm, in which a reward model trained on human preferences guides PPO-based policy optimization~\cite{ouyang2022training,schulman2017proximal}. DPO~\cite{rafailov2023direct} later simplified this by eliminating the reward model, but its formulation is tied to pairwise preference data. It does not accommodate verifiable rewards such as mathematical correctness or code execution results, which are deterministically computed from model outputs. This distinction motivates a separate line of work on reward-driven RL, where the policy is optimized directly against task-specific, automatically evaluated reward functions.

Mathematical reasoning is a natural domain for such methods. STaR~\cite{zelikman2022star} bootstrapped reasoning via iterative self-training on correct rationales. DeepSeekMath~\cite{shao2024deepseekmath} introduced GRPO and DeepSeek-R1~\cite{deepseek2025r1} later demonstrated that large-scale RL with a simple rule-based reward can elicit sophisticated reasoning behaviors. DAPO~\cite{yu2025dapo} proposed a Dynamic sAmpling Policy Optimization algorithm and introduced four key techniques to stabilize RL in the long chain-of-thought setting, including Clip-Higher for wider exploration and dynamic sampling to filter out low-quality prompts. GSPO~\cite{zheng2025gspo} and GMPO~\cite{zhao2025gmpo}, the two variants directly evaluated in this report, address sequence-level versus token-level optimization and arithmetic-mean versus geometric-mean aggregation, respectively.

\subsection{Chain-of-Thought Reasoning}
Chain-of-thought (CoT) prompting~\cite{wei2022chain} elicits step-by-step reasoning from LLMs by augmenting the input with intermediate reasoning traces, substantially improving performance on multi-step arithmetic, and commonsense tasks. Kojima et al.~\cite{kojima2022zero} later showed that the same effect can be achieved without exemplars, simply through the prompt ``Let's think step by step'', suggesting that structured reasoning is an inherent capability that can be triggered rather than taught. Self-consistency~\cite{wang2023self} further improves reliability by sampling multiple reasoning paths and marginalizing over them via majority voting. This exploits the observation that correct reasoning tends to be more consistent across diverse samples than incorrect reasoning.

Beyond prompting, CoT data has become central to post-training. Large-scale datasets such as MetaMathQA~\cite{yu2023metamath}, OpenMathInstruct-1~\cite{toshniwal2024openmathinstruct}, and NuminaMath-CoT~\cite{numina2024math} provide large collections of problem--solution pairs with CoT-format rationales, enabling SFT to instill structured step-by-step reasoning as a behavioral prior. In this report, we further distill wireless-specific CoT trajectories from DeepSeek-V3 to align this reasoning prior with the target signal-processing domain. In the RL stage, this prior shapes the model's exploration, since responses already follow a derivation-then-answer format and the verifiable reward can focus on correctness rather than format discovery. The interaction between this CoT initialization and subsequent RL is a key axis of the present study.

\subsection{LLMs in Signal Reasoning}
Signal processing and wireless communications impose stringent requirements on mathematical precision, particularly for tasks such as channel estimation, interference management, and beamforming~\cite{cadambe2008interference,gesbert2010multi}. Some preliminary works have explored the use of LLMs in wireless contexts, focusing on domain-specific knowledge extraction and basic recall of technical standards~\cite{maatouk2023large,maatouk2024telellms,colle2025telemath,tong2026wirelessbench}. Notably, TelecomGPT~\cite{zou2024telecomgpt} has extended LLM capabilities to higher-level tasks such as wireless-specific code generation and formula completion. However, these early works primarily emphasize knowledge retrieval or summarization, without systematically evaluating whether LLMs can perform the multi-step mathematical reasoning required in actual signal processing engineering systems.

WirelessMathBench~\cite{li2025wirelessmathbench} took a step by introducing a curated benchmark of 587 expert-level signal processing math problems. Its successor \bench{}~\cite{li2025wirelessmathlm} scaled the dataset to 4,027 problems and demonstrated that compact models fine-tuned with GRPO can approach the performance of much larger general-purpose models. Building on this line of work, the present report systematically compares several reinforcement fine-tuning strategies for adapting a sub-10B model to signal-domain mathematical reasoning.

\section{Method}
\subsection{Supervised Fine-Tuning}
Given a supervised dataset $\mathcal{D}_{\mathrm{SFT}}=\{(x,y)\}$, SFT minimizes the autoregressive negative log-likelihood:
\begin{equation}
  \mathcal{L}_{\mathrm{SFT}}(\theta)
  = - \mathbb{E}_{(x,y)\sim\mathcal{D}_{\mathrm{SFT}}}
  \frac{1}{|y|}\sum_{t=1}^{|y|}\log \pi_\theta(y_t\mid x,y_{<t}).
\end{equation}
In this project, SFT is applied only in Route B using \sftdata{}. The purpose is to initialize the model with both signal-domain reasoning trajectories and general mathematical CoT behavior: step decomposition, symbolic manipulation, and final-answer formatting.

\subsection{Group Relative Policy Optimization}
Proximal Policy Optimization (PPO)~\cite{schulman2017proximal} has become the dominant RL algorithm for LLM post-training, owing to its stable clipped objective and compatibility with the autoregressive token-generation paradigm. 
However, PPO relies on a separate critic network to estimate per-token state values, which are then combined with empirical returns via Generalized Advantage Estimation (GAE) to produce low-variance advantage signals. 
Training the critic alongside the policy approximately doubles GPU memory consumption and inaccurate critic can bias the advantage and misdirect policy updates.

GRPO, proposed in DeepSeekMath~\cite{shao2024deepseekmath} and later adopted in DeepSeek-R1~\cite{deepseek2025r1}, sidesteps the critic entirely. For each prompt $x$, GRPO samples a group of $G$ responses $\{y_i\}_{i=1}^{G}$ from the old policy $\pi_{\theta_{\mathrm{old}}}$ and normalizes the reward within the group to obtain a baseline-free advantage:
\begin{equation}
  \widehat{A}_i = \frac{r_i - \mathrm{mean}(\{r_j\}_{j=1}^{G})}
  {\mathrm{std}(\{r_j\}_{j=1}^{G}) + \epsilon}.
\end{equation}

\begin{figure}[tbp]
  \centering
  \includegraphics[width=0.9\textwidth]{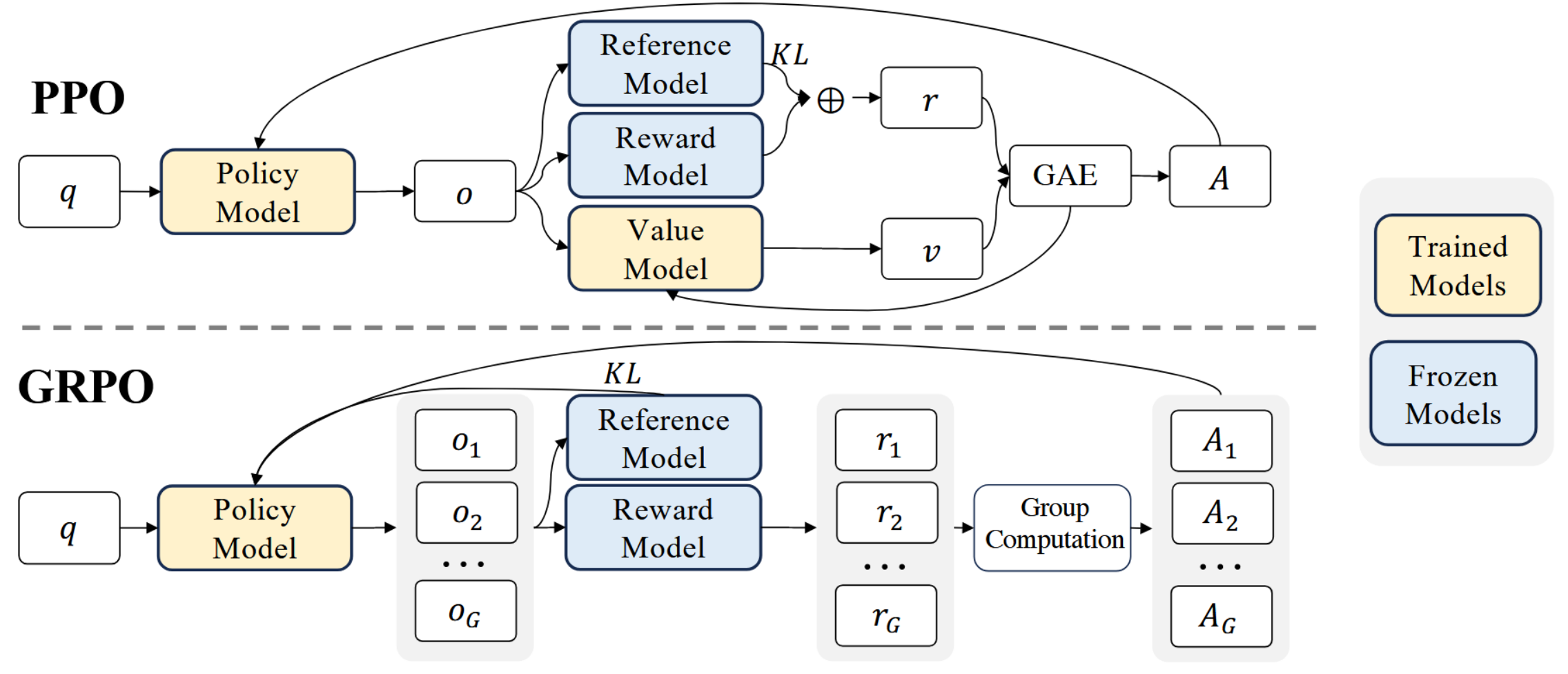}
  \caption{Comparison of PPO and GRPO architectures, adapted from DeepSeek-R1~\cite{deepseek2025r1}. PPO requires a separate critic network for advantage estimation via GAE, while GRPO eliminates the critic by computing group-relative advantages from multiple sampled responses.}
  \label{fig:ppo-vs-grpo}
\end{figure}

GRPO maximizes a clipped token-level surrogate with a KL penalty that regularizes the current policy towards a frozen reference $\pi_{\mathrm{ref}}$:
\begin{equation}
  \mathcal{J}_{\mathrm{GRPO}}(\theta)
  =
  \mathbb{E}
  \left[
  \frac{1}{G}\sum_{i=1}^{G}
  \frac{1}{|y_i|}\sum_{t=1}^{|y_i|}
  \min\left(
  \rho_{i,t}(\theta)\widehat{A}_i,
  \mathrm{clip}(\rho_{i,t}(\theta),1-\varepsilon,1+\varepsilon)\widehat{A}_i
  \right)
  - \beta\,D_{\mathrm{KL}}(\pi_\theta \| \pi_{\mathrm{ref}})
  \right],
\end{equation}
where $\rho_{i,t}(\theta)=\pi_\theta(y_{i,t}\mid x,y_{i,<t})/
\pi_{\theta_{\mathrm{old}}}(y_{i,t}\mid x,y_{i,<t})$ is the token-level importance ratio and $\beta$ controls the KL penalty strength. The KL divergence is estimated by the unbiased, guaranteed non-negative estimator
\begin{equation}
  D_{\mathrm{KL}}^{(t)}
  =
  \frac{\pi_{\mathrm{ref}}(y_{i,t}\mid x,y_{i,<t})}
  {\pi_\theta(y_{i,t}\mid x,y_{i,<t})}
  - \log\frac{\pi_{\mathrm{ref}}(y_{i,t}\mid x,y_{i,<t})}
  {\pi_\theta(y_{i,t}\mid x,y_{i,<t})}
  - 1.
\end{equation}
All tokens in a response share the same advantage $\widehat{A}_i$, since only the final token typically receives a reward.
This design offers three key benefits. It halves memory usage by removing the critic, which is essential when training on consumer or mid-range GPUs. The group-relative normalization is invariant to the scale and shift of the reward function, so the same clipping hyperparameters transfer across tasks with different reward ranges without manual tuning. And the group mean acts as an implicit baseline, providing variance reduction comparable to a learned value function at no extra cost. These properties make GRPO particularly well-suited for reasoning benchmarks like \bench{}, where a clean verifiable reward signal is available for each complete response.

\subsection{Group Sequence Policy Optimization}
In large-scale RL training, a big rollout batch is typically partitioned into mini-batches for gradient updates to maximize hardware utilization. This introduces an off-policy setting where responses are sampled from an old policy $\pi_{\theta_{\mathrm{old}}}$ rather than the current policy $\pi_\theta$, which is why clipping mechanisms are employed in PPO and GRPO. However, GSPO identifies a more fundamental issue in GRPO: its token-level importance sampling is ill-posed. Importance sampling estimates an expectation under a target distribution by averaging re-weighted samples drawn from a behavior distribution, and the correction relies on aggregating over multiple samples from the same distribution. GRPO applies the importance weight $\rho_{i,t}(\theta)$ at each token based on a single draw from the next-token distribution, which fails to perform meaningful distribution correction and instead injects high-variance noise into the gradient. This noise accumulates over long sequences and can lead to irreversible model collapse.

The root cause points to a simple principle that the unit of optimization should match the unit of reward. Since the reward is assigned to an entire response, applying off-policy correction and clipping at the token level is fundamentally mismatched. GSPO addresses this by shifting the optimization unit from individual tokens to complete sequences. The sequence-level importance ratio is length-normalized to unify the numerical range across responses of different lengths:
\begin{equation}
  s_i(\theta)
  =
  \left[
  \frac{\pi_\theta(y_i\mid x)}
  {\pi_{\theta_{\mathrm{old}}}(y_i\mid x)}
  \right]^{1/|y_i|}
  =
  \exp\left(
  \frac{1}{|y_i|}\sum_{t=1}^{|y_i|}
  \log \rho_{i,t}(\theta)
  \right).
\end{equation}
The clipping are then applied at the sequence level:
\begin{equation}
  \mathcal{J}_{\mathrm{GSPO}}(\theta)
  =
  \mathbb{E}
  \left[
  \frac{1}{G}\sum_{i=1}^{G}
  \min\left(
  s_i(\theta)\widehat{A}_i,
  \mathrm{clip}(s_i(\theta),1-\varepsilon,1+\varepsilon)\widehat{A}_i
  \right)
  \right].
\end{equation}
By operating at the sequence level, GSPO aligns the optimization unit with the reward unit and mitigates the training collapse caused by token-level importance sampling noise.

\subsection{Geometric-Mean Policy Optimization}

\begin{figure}[tbp]
  \centering
  \includegraphics[width=0.95\textwidth]{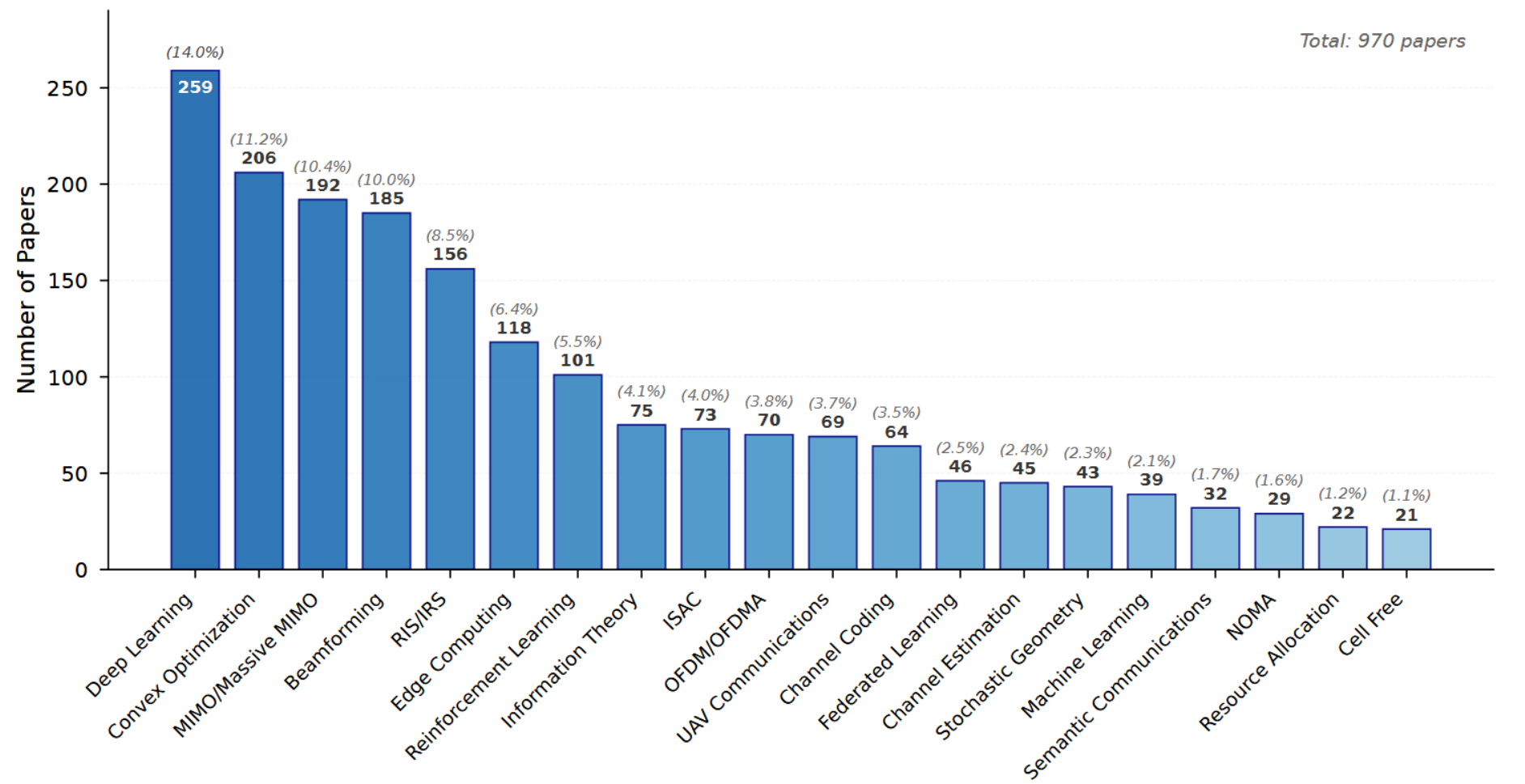}
  \caption{Distribution of the top 20 key techniques across the 970 source papers in \bench{}~\cite{li2025wirelessmathlm}.}
  \label{fig:paper-distribution}
\end{figure}

In GRPO, the per-token importance ratios $\rho_{i,t}(\theta)$ are averaged arithmetically across the response. While simple, this arithmetic mean is highly sensitive to outliers: a single token with an extreme ratio (e.g., where the current policy assigns dramatically higher or lower probability than the old policy) can dominate the entire average, especially in long responses. GMPO addresses this by replacing the arithmetic mean with a geometric mean. By the AM--GM inequality, $|\mathcal{J}_{\mathrm{GMPO}}| \leq |\mathcal{J}_{\mathrm{GRPO}}|$, which gives GMPO a strictly narrower value range and lower optimization variance. The geometric mean effectively dampens the influence of any single token, requiring a more uniform consensus across the sequence for the update to be large.

A further difference from GSPO concerns the clipping granularity. GSPO clips at the sequence level, which can discard the entire gradient signal for a response when its sequence-level ratio exceeds the clipping threshold. GMPO instead clips individual token-level ratios in log-space before geometric aggregation. This preserves gradient signals for unclipped tokens while only suppressing the extreme ones, making more efficient use of the sampled data. Inspired by DAPO's Clip-Higher strategy~\cite{yu2025dapo}, GMPO adopts a wider clipping range in the log-ratio space, i.e., $\rho_{i,t}^{\mathrm{sgn}(\widehat{A}_i)}\in[e^{-0.4}, e^{0.4}]$, to improve exploration while maintaining stable optimization. The simplified (unclipped) GMPO objective is
\begin{equation}
  \mathcal{J}_{\mathrm{GMPO}}(\theta)
  =
  \mathbb{E}
  \left[
  \frac{1}{G}\sum_{i=1}^{G}
  \left(
  \prod_{t=1}^{|y_i|}
  \frac{\pi_\theta(y_{i,t}\mid x,y_{i,<t})}
  {\pi_{\theta_{\mathrm{old}}}(y_{i,t}\mid x,y_{i,<t})}
  \middle| \widehat{A}_i \middle|
  \right)^{\!\frac{1}{|y_i|}}
  \mathrm{sgn}(\widehat{A}_i)
  \right].
\end{equation}
For the full objective, token-level clipping is applied:
\begin{equation}
  \mathcal{J}_{\mathrm{GMPO}}(\theta)
  =
  \mathbb{E}
  \left[
  \frac{1}{G}\sum_{i=1}^{G}
  \left\{
  \prod_{t=1}^{|y_i|}
  \min\!\Big(
  \rho_{i,t}(\theta)\widehat{A}_i,\;
  \mathrm{clip}\big(
  \rho_{i,t}(\theta),\,
  \varepsilon_{\mathrm{low}},\,
  \varepsilon_{\mathrm{high}}
  \big)\widehat{A}_i
  \Big)
  \right\}^{\!\frac{1}{|y_i|}}
  \cdot
  \mathrm{sgn}(\widehat{A}_i)
  \right],
\end{equation}
where $\mathrm{sgn}(\widehat{A}_i)$ returns $+1$ for positive advantages and $-1$ otherwise, and $(\varepsilon_{\mathrm{low}},\varepsilon_{\mathrm{high}})$ are the lower and upper clipping thresholds on the importance ratio (e.g., $e^{-0.4}$ and $e^{0.4}$, following DAPO's Clip-Higher strategy). Unlike GSPO, which clips at the sequence level, GMPO clips individual token-level ratios, preserving gradient signals for unclipped tokens while suppressing extreme outliers. In our experiments, GMPO is evaluated as a stability-oriented alternative to GRPO under the same prompts, reward function, and base model.

\section{Task and Training Routes}
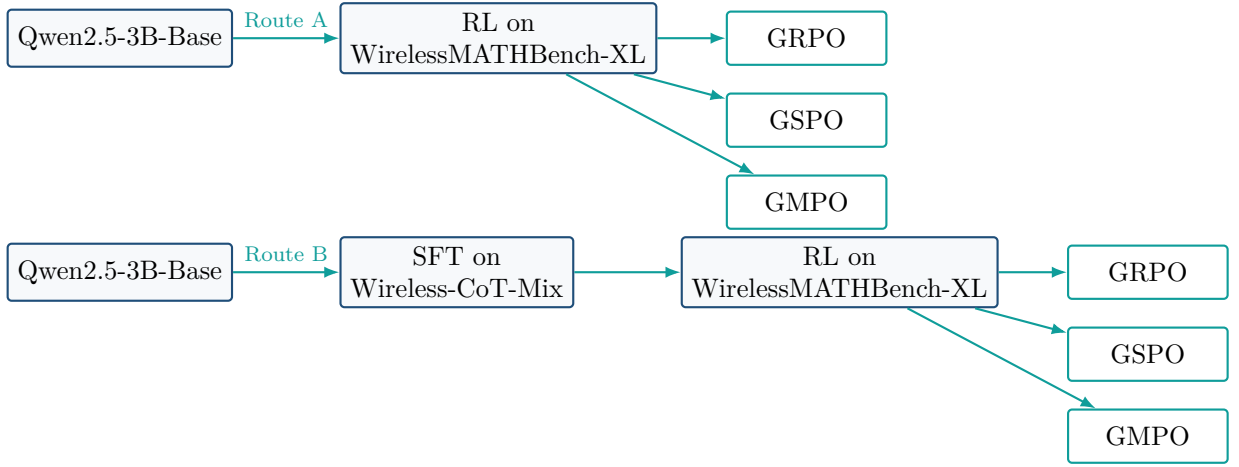
\begin{figure}[tbp]
  \centering
  \resizebox{\textwidth}{!}{%
  \begin{tikzpicture}[
    node distance=0.85cm,
    box/.style={draw=paperblue, thick, rounded corners=1.5pt, fill=lightbg, minimum width=2.9cm, minimum height=0.75cm, align=center, font=\small},
    alg/.style={draw=accent, thick, rounded corners=1.5pt, fill=white, minimum width=2.1cm, minimum height=0.7cm, align=center, font=\small},
    arr/.style={-{Latex[length=2mm]}, thick, color=accent}
  ]
    \node[box] (basea) {\model};
    \node[box, right=1.4cm of basea] (rla) {RL on\\\bench{}};
    \node[alg, right=0.9cm of rla] (grpoa) {GRPO};
    \node[alg, below=0.35cm of grpoa] (gspoa) {GSPO};
    \node[alg, below=0.35cm of gspoa] (gmpoa) {GMPO};
    \draw[arr] (basea) -- node[above,font=\scriptsize]{Route A} (rla);
    \draw[arr] (rla) -- (grpoa);
    \draw[arr] (rla) -- (gspoa);
    \draw[arr] (rla) -- (gmpoa);

    \node[box, below=2.3cm of basea] (baseb) {\model};
    \node[box, right=1.4cm of baseb] (sft) {SFT on\\\sftdata{}};
    \node[box, right=1.4cm of sft] (rlb) {RL on\\\bench{}};
    \node[alg, right=0.9cm of rlb] (grpob) {GRPO};
    \node[alg, below=0.35cm of grpob] (gspob) {GSPO};
    \node[alg, below=0.35cm of gspob] (gmpob) {GMPO};
    \draw[arr] (baseb) -- node[above,font=\scriptsize]{Route B} (sft);
    \draw[arr] (sft) -- (rlb);
    \draw[arr] (rlb) -- (grpob);
    \draw[arr] (rlb) -- (gspob);
    \draw[arr] (rlb) -- (gmpob);
  \end{tikzpicture}
  }
  \caption{Two-route training design. Route A applies RL directly to \model{}. Route B first performs SFT on \sftdata{} and then applies signal-domain RL.}
  \label{fig:routes}
\end{figure}
\subsection{Benchmark Task and Dataset}
\bench{}~\cite{li2025wirelessmathlm} contains 4,027 graduate-level signal processing mathematical problems curated from 970 technical papers. The dataset spans a broad range of topics, including MIMO, RIS, ISAC, beamforming, NOMA, satellite communications, and so on. The problems are organized into two question types: multiple-choice questions and fill-in-the-blank questions. The dataset is split into 3,227 training examples and 800 test examples.
Each sample contains a prompt $x$ assembled from structured fields, including a technical background passage, a question text, an optional equation, and optional answer choices, and a gold answer $a$. The model generates a response $y$ that should contain a concise reasoning process and a final answer in \verb|\boxed{}|.

\sftdata{} is the SFT corpus constructed for Route~B. It contains 3,542 examples, including 1,546 wireless-domain question--reasoning pairs and 1,996 general-domain examples sampled from NuminaMath-CoT~\cite{numina2024math}. The wireless portion is distilled from DeepSeek-V3 using two sources: \bench{}, whose official split contains 3,227 training and 800 test instances, and WirelessMathBench~\cite{li2025wirelessmathbench}, which contains 587 instances. For \bench{}, we sample approximately 50\% of the training split for SFT construction, corresponding to about 1,613 candidate problems, and obtain 1,503 verified CoT trajectories after generation and answer checking. For WirelessMathBench, the same pipeline produces 483 verified CoT trajectories. After length filtering, the final wireless SFT subset contains 1,546 usable question--reasoning pairs.

\begin{table}[tbp]
  \centering
  \caption{Main experimental methods.}
  \label{tab:systems}
  \begin{tabular}{lccc}
    \toprule
    ID & Route & Initialization & RL Algorithm \\
    \midrule
    Base & / & Base & / \\
    A-GRPO & RL & Base & GRPO \\
    A-GSPO & RL & Base & GSPO \\
    A-GMPO & RL & Base & GMPO \\
    Base-SFT & SFT & Base + SFT & / \\
    B-GRPO & SFT + RL & Base + SFT & GRPO \\
    B-GSPO & SFT + RL & Base + SFT & GSPO \\
    B-GMPO & SFT + RL & Base + SFT & GMPO \\
    \bottomrule
  \end{tabular}
\end{table}
\subsection{Route A: Direct RL from \model{}}
As shown in Figure~\ref{fig:routes}, the first route starts from the raw \model{} checkpoint and directly performs RL on \bench{}. This route is designed to answer the following question:
\begin{quote}
Can signal-domain verifiable rewards alone induce useful mathematical reasoning behavior in a compact base model?
\end{quote}
Under this route, we compare three group-based RL algorithms, namely GRPO, GSPO, and GMPO, to evaluate their effectiveness when applied directly to a base model without any SFT initialization.

\subsection{Route B: SFT on \sftdata{} Followed by RL}
As shown in Figure~\ref{fig:routes}, the second route first fine-tunes \model{} on \sftdata{}, then applies RL on \bench{}. This route is designed to answer the following question:
\begin{quote}
Does domain-aware CoT SFT provide a better initialization for subsequent signal-domain RL?
\end{quote}
The SFT stage is intended to initialize the model with wireless-domain reasoning patterns, mathematical decomposition habits, and final-answer discipline while preserving general mathematical reasoning through the NuminaMath-CoT mixture. The subsequent RL stage then further specializes this prior toward signal-processing tasks with verifiable rewards. As in Route A, we compare GRPO, GSPO, and GMPO at the RL stage.

\section{Experimental Setup}

\subsection{Compared Methods}
The eight compared methods are listed in \Cref{tab:systems}, comprising two baselines (Base and Base-SFT) and six RL variants (A-GRPO, A-GSPO, A-GMPO, B-GRPO, B-GSPO, B-GMPO). This design isolates two questions: whether SFT improves the RL starting point, and which RL algorithm is most effective under each starting point.

\subsection{Hyperparameter Settings}
All experiments are conducted using the open-source VeRL framework~\footnote{\url{https://github.com/volcengine/verl}}. The SFT stage trains \model{} on \sftdata{} for 1 epoch with a learning rate of $3\times10^{-6}$ and a batch size of 36. The RL stage uses a group size of 4, a learning rate of $1\times10^{-6}$, and a batch size of 18, training for 1,000 steps on three NVIDIA RTX 4090 GPUs. The maximum response length is set to 768 tokens for both SFT training and RL generation. Detailed hyperparameters are provided in \Cref{tab:config} in the appendix.

\subsection{Reward Design}
The reward function extracts all \verb|\boxed{}| expressions from the model's response $y$ and compares them element-wise against the ground-truth answer $a$. Let $M$ denote the number of ground-truth boxed expressions and $m$ the number of extracted expressions that match the ground truth exactly. The reward is defined as

\begin{equation}
  R(y,a) =
  \begin{cases}
    0.0, & \text{no \texttt{\string\boxed{}} detected},\\[2pt]
    0.1, & m=0,\\[2pt]
    0.2, & 0 < m < M,\\[2pt]
    1.0, & m = M.
  \end{cases}
\end{equation}
\begin{table}[tbp]
  \centering
  \caption{Performance of SignalReasoner models on the \bench{} test set (\%). MCQ: Multiple Choice Questions, Fill-in: Fill-in-the-blank, FEC: Full Equation Completion.}
  \label{tab:main-results}
  \small
  \begin{tabular}{lcccccc}
    \toprule
    Model & Size & MCQ & Fill-in & FEC & Overall & Output Tokens \\
    \midrule
    \multicolumn{7}{l}{\textit{SignalReasoner 3B Models}} \\
    Base & 3B & 28.57 & 9.03 & 9.42 & 12.37 & 907.41 \\
    Base-SFT & 3B & 42.11 & 26.68 & 24.61 & 28.75 & 604.45 \\
    A-GRPO & 3B & 57.89 & 30.88 & 27.23 & 34.50 & 668.33 \\
    A-GSPO & 3B & 61.65 & 30.88 & 26.18 & 34.88 & 279.25 \\
    A-GMPO & 3B & 51.88 & 32.98 & 26.70 & 34.63 & 395.84 \\
    B-GRPO & 3B & 49.62 & 33.61 & 25.65 & 34.38 & 600.10 \\
    B-GSPO & 3B & 57.89 & 35.08 & 30.37 & 37.75 & \textbf{243.59} \\
    B-GMPO & 3B & 54.14 & 38.03 & 31.41 & 39.12 & 258.87 \\
    \bottomrule
  \end{tabular}
\end{table}

\section{Experimental Results}
\subsection{Main Results}
\Cref{tab:main-results} reports the main comparison. RL training consistently improves performance, with overall accuracy rising from 12.37\% for the Base model to 34--39\% after reinforcement fine-tuning.
In addition, SFT alone on \sftdata{} substantially improves overall accuracy to 28.75\%, showing that the distilled wireless-domain CoT data injects useful signal-specific reasoning behavior before RL. 
SFT also reduces the average output length from 907.41 to 604.45 tokens, suggesting that the mixed CoT corpus teaches a more structured and concise response format.
Meanwhile, Route~B achieves the highest overall accuracy. B-GMPO reaches 39.12\% and B-GSPO reaches 37.75\%, suggesting that SFT initialization provides a clear benefit when combined with sequence-level or geometric-mean optimization. 
In contrast, for GRPO, direct RL and the SFT-pretrained variant remain close, implying that GRPO's token-level advantage estimation may be less sensitive to the quality of the initial policy.

Meanwhile, a striking pattern emerges in output token length. GRPO variants produce 545--668 tokens on average, maintaining substantial reasoning traces. 
In contrast, GSPO and GMPO drastically reduce output length. A-GSPO and A-GMPO produce 279.25 and 395.84 tokens respectively, while B-GSPO and B-GMPO further reduce the average length to 243.59 and 258.87 tokens. 
At this token budget, the model can still provide concise derivations, but the reasoning traces are much shorter than those produced by GRPO.
We attribute this phenomenon to two compounding effects.
First, both GSPO and GMPO implicitly penalize long sequences. 
GSPO computes a single importance ratio per sequence and applies the same update magnitude to every token regardless of length. 
A long incorrect response, which is common during early exploration, therefore receives the same per-sequence penalty as a short incorrect one but distributes that penalty across more tokens, resulting in a stronger cumulative suppression of long outputs over training. 
GMPO's geometric-mean objective suppresses outlier token-level ratios, which disproportionately arise in long reasoning chains with high variance, thereby reducing the effective update signal for lengthy responses. 
Second, SFT pre-training on \sftdata{} already biases the model toward a structured but concise output format. When this compact prior is combined with the implicit length penalty of GSPO and GMPO, the RL optimization may discover a shortcut, namely outputting the answer with little reasoning, since the verifiable reward only checks the final \verb|\boxed{}| expression and does not reward intermediate steps. 
The result is a ``reasoning shortening'' effect in which the model achieves competitive or even superior accuracy with dramatically fewer tokens, at the cost of less detailed step-by-step derivations. 
Whether this shortening is desirable depends on the deployment context. 
It improves inference efficiency but eliminates the auditability and error-diagnosis benefits of explicit reasoning traces.

\begin{figure}[tbp]
  \centering
  \includegraphics[width=0.9\textwidth]{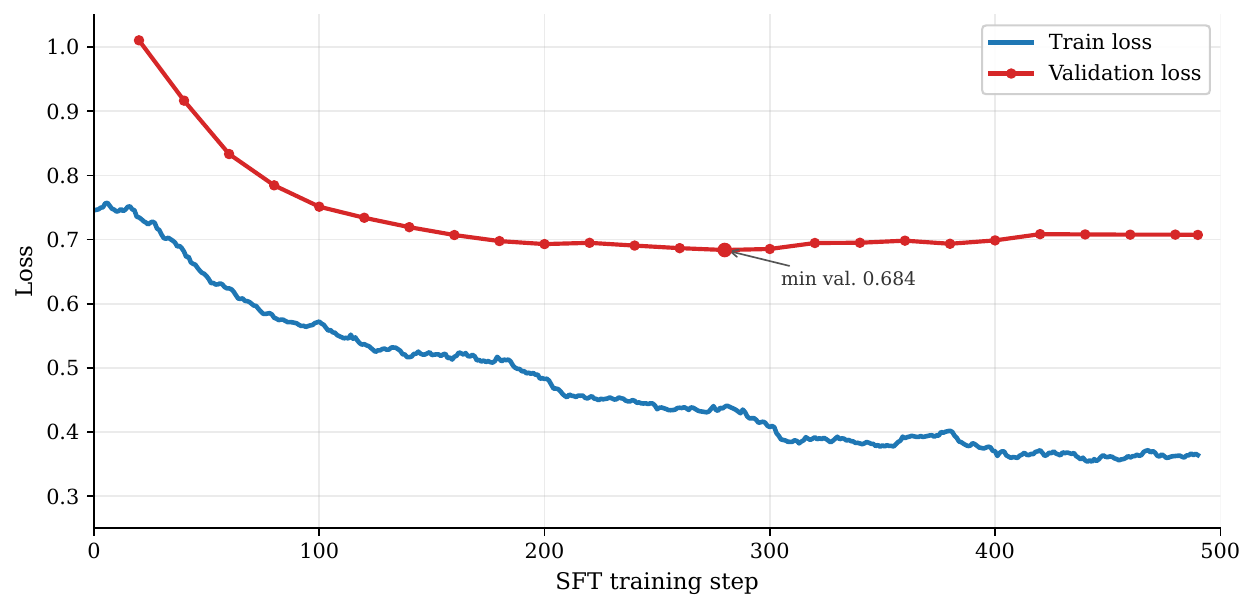}
  \caption{SFT loss curves on \sftdata{}. The blue curve shows smoothed training loss, and the red curve shows validation loss evaluated every 20 steps.}
  \label{fig:sft-loss}
\end{figure}

\subsection{SFT Training Dynamics}
\Cref{fig:sft-loss} shows the training and validation loss curves during the SFT stage on \sftdata{}. The smoothed training loss decreases from roughly 0.75 in the early steps to about 0.36 by the end of training, indicating stable optimization on the mixed wireless and general-domain CoT corpus. 
The validation loss drops rapidly from 1.01 to a minimum of approximately 0.684, after which it plateaus and slightly increases to around 0.71. 
This pattern suggests that the model quickly learns the supervised CoT format and domain-specific answer structure, while later updates mainly continue fitting the training distribution rather than improving validation loss. 
Empirically, however, we found that using the final checkpoint as the initialization for RL yields noticeably worse downstream accuracy compared to an earlier checkpoint. 
We attribute this to over-training on the SFT distribution. 
By the end of training, the model's output distribution can become overly concentrated on the specific formatting and answer patterns of \sftdata{}, reducing policy entropy and limiting the exploration capacity needed for RL to adapt to the signal domain. 
The excessively narrow prior effectively traps the RL optimizer in a local region of the policy space from which it is difficult to escape. 
Based on this observation, all Route~B experiments in this report use an intermediate SFT checkpoint, which balances structured reasoning capability with sufficient policy diversity for effective RL fine-tuning.

\begin{figure}[tbp]
  \centering
  \includegraphics[width=0.96\textwidth]{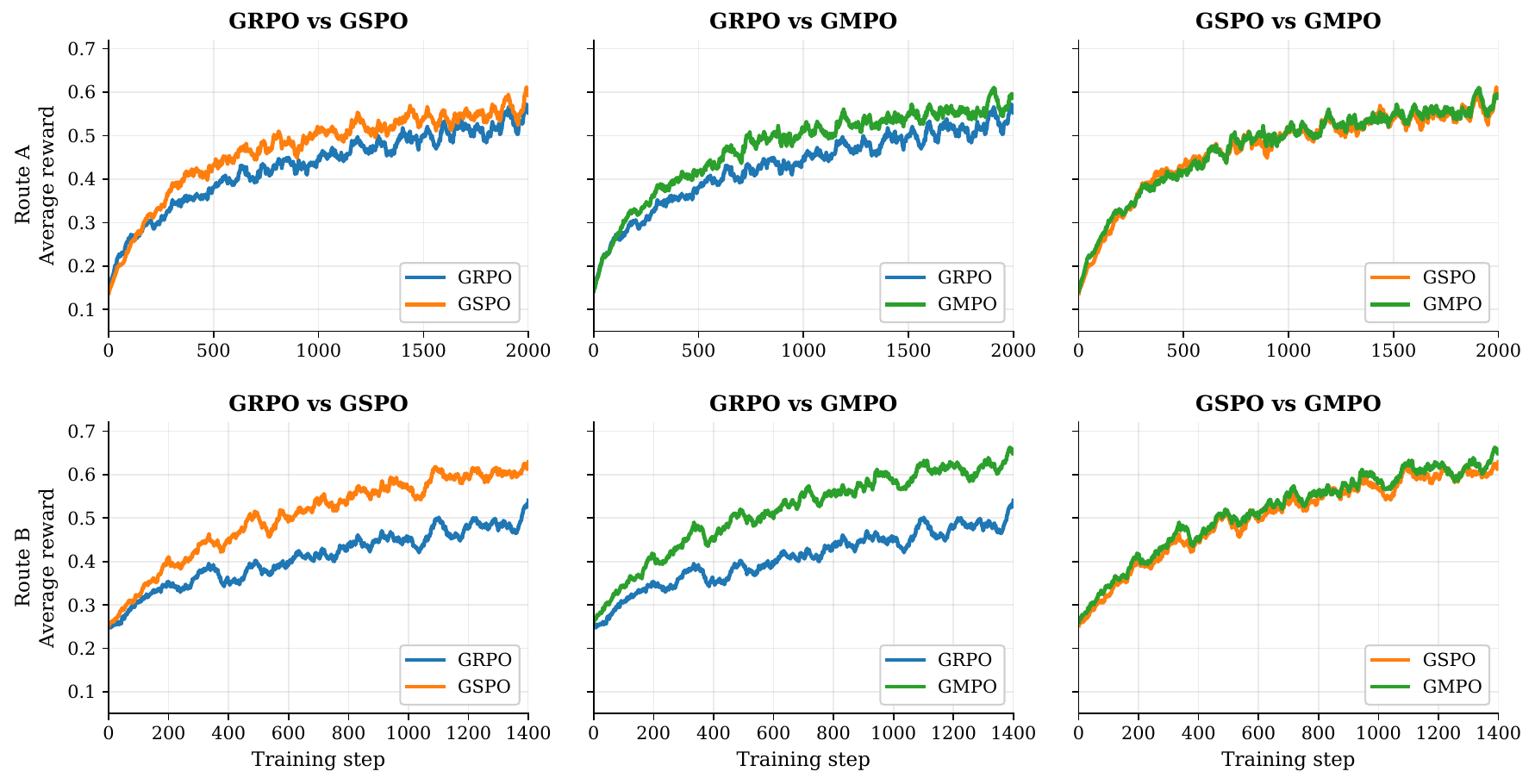}
  \caption{Algorithm comparisons within each route. Top row: direct RL from the Base model (Route~A), truncated at 2,000 steps. Bottom row: the SFT+RL route, where RL starts after supervised training on \sftdata{} (Route~B), truncated at 1,400 steps.}
  \label{fig:rl-algo-compare}
\end{figure}

\subsection{RL Training Dynamics}
\begin{figure}[htbp]
  \centering
  \includegraphics[width=0.96\textwidth]{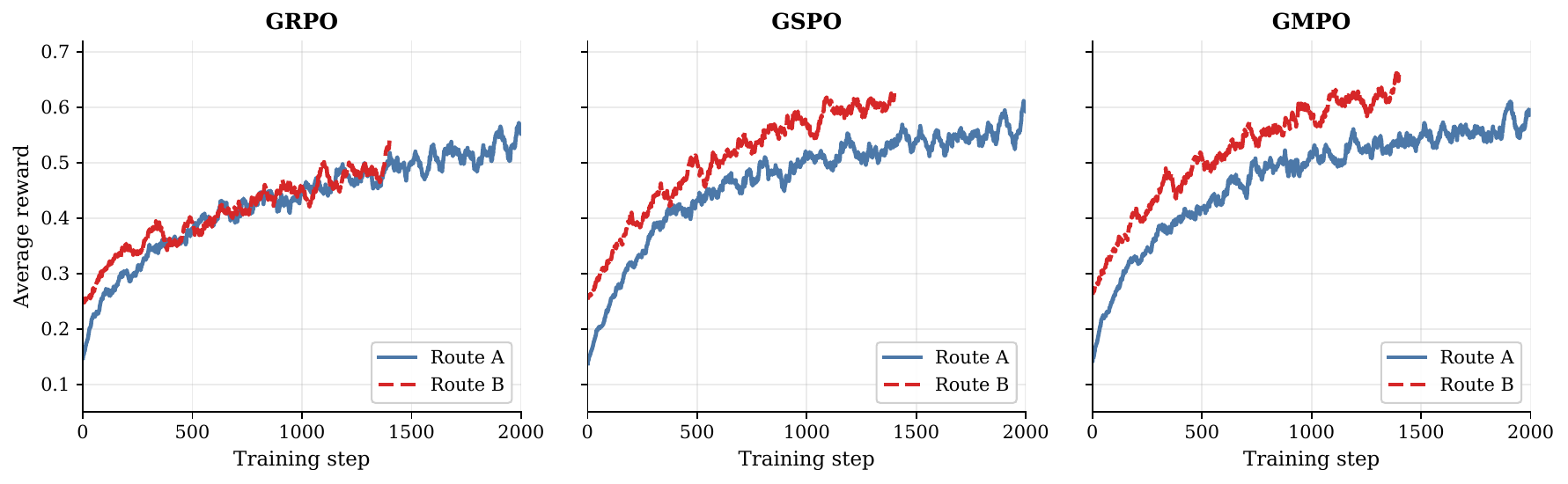}
  \caption{Route comparisons per algorithm. Solid lines denote Route~A, direct RL from the Base model, truncated at 2,000 steps. Dashed lines denote Route~B, the SFT+RL route, truncated at 1,400 steps.}
  \label{fig:rl-route-compare}
\end{figure}

\Cref{fig:rl-algo-compare} and \Cref{fig:rl-route-compare} show the reward curves for all six RL variants over the available training steps.
For the SFT+RL route, the plotted trajectories correspond to RL runs that start from the intermediate \sftdata{} checkpoint described above, rather than from the raw Base model. Thus, the bottom row of \Cref{fig:rl-algo-compare} should be read as the second-stage optimization behavior after supervised CoT distillation: SFT first establishes a wireless-aware reasoning prior, and RL then refines the policy with the same verifiable reward used in Route~A. 
\Cref{fig:rl-algo-compare} compares algorithm pairs under the same initialization. 
Under the Base route, GSPO and GMPO exhibit more stable reward improvement than GRPO, which shows larger oscillations particularly in the early phase. 
GSPO's sequence-level updates produce the smoothest curves, consistent with its design of reducing per-token variance. 
Under the SFT route, the gap between algorithms narrows. 
GSPO and GMPO still maintain slightly lower variance than GRPO. 
Notably, GMPO achieves the highest final reward under both routes, corroborating its top overall accuracy in \Cref{tab:main-results}. Moreover, both GSPO and GMPO exhibit markedly faster convergence than GRPO. Their reward curves rise more steeply in the first few hundred steps and reach a higher plateau earlier, indicating superior training efficiency.

\Cref{fig:rl-route-compare} compares Route~A and Route~B for each algorithm. 
For GRPO, the B-GRPO shows a modest advantage in early-step reward growth but the two routes converge to similar levels by the end of training, consistent with the small accuracy gap observed in \Cref{tab:main-results}. 
For GSPO and GMPO, the SFT-initialized variants achieve higher reward levels more rapidly than their Base counterparts, indicating that the structured CoT prior from SFT interacts synergistically with sequence-level and geometric-mean optimization. 
This acceleration is most pronounced for GMPO, where B-GMPO consistently leads A-GMPO throughout training and ultimately delivers the highest accuracy.

\section{Conclusion}
This report studied reinforcement fine-tuning of \model{} for signal mathematical reasoning on \bench{}, comparing direct RL with an SFT-then-RL route under GRPO, GSPO, and GMPO. The results show that both stages contribute: SFT alone improves overall accuracy from 12.37\% for the Base model to 28.75\%, while subsequent RL reaches 39.12\% with B-GMPO. The SFT-then-RL route is particularly effective for GSPO and GMPO, with B-GSPO and B-GMPO achieving 37.75\% and 39.12\%, respectively. The advantage is less pronounced for GRPO, whose direct-RL and SFT-then-RL variants obtain similar overall scores. Thus, SFT provides a useful domain-specific starting point, but its benefit depends on the RL objective.

Across the three algorithms, GSPO and GMPO show the clearest gains over GRPO in the SFT-then-RL setting, together with faster early reward growth. They also produce much shorter responses: B-GSPO and B-GMPO average 243.59 and 258.87 output tokens, compared with 600.10 for B-GRPO. This shortening improves output efficiency, but it should not be interpreted as evidence that the model generates more complete reasoning. Because the current verifiable reward primarily checks the final boxed answer, optimization can favor concise responses with limited intermediate derivation. The findings therefore expose a central limitation of answer-only rewards for small reasoning models: accuracy and efficiency can improve while the transparency of the reasoning trace decreases. Future work should combine answer verification with process-sensitive rewards or auxiliary supervision that explicitly preserves useful intermediate reasoning.
\newpage
\printbibliography

\newpage
\appendix
\setcounter{table}{0}
\renewcommand{\thetable}{A\arabic{table}}
\setcounter{figure}{0}
\renewcommand{\thefigure}{A\arabic{figure}}
\begin{center}
{\Large\bfseries Appendix}
\end{center}
\vspace{0.8em}

\section{Data and Model Preprocessing}

\subsection{SFT Data}
The SFT stage uses \sftdata{}, a 3,542-example corpus constructed from both wireless-domain and general-domain mathematical reasoning data. The wireless portion is obtained by distilling CoT trajectories from DeepSeek-V3. We use two sources: \bench{}, which contains 3,227 training and 800 test instances, and WirelessMathBench, which contains 587 instances. For \bench{}, we sample approximately half of the training split for SFT construction, yielding about 1,613 candidate problems. The distillation and verification pipeline produces 1,503 valid CoT trajectories from this subset and 483 valid CoT trajectories from WirelessMathBench.

For each candidate problem, the prompt includes the domain background, question, masked equation, and ground-truth answer. DeepSeek-V3 is asked to generate concise step-by-step reasoning that naturally reaches the provided answer and ends with the final result enclosed in \verb|\boxed{}|. We then verify that every generated \verb|\boxed{}| expression exactly matches the corresponding ground-truth answer. Failed generations are retried up to five times and discarded if they still fail verification. The generated reasoning is restricted to 512 tokens, and the sampling temperature is set to 1.0.

After answer verification and length filtering, the wireless SFT subset contains 1,546 question--reasoning pairs. To preserve general mathematical reasoning ability, we add 1,996 examples sampled from NuminaMath-CoT~\cite{numina2024math}, yielding the final 3,542-example SFT corpus. Each instance is reformatted as a two-turn conversation: a user message containing the problem plus a ``reason step by step'' instruction, and an assistant message containing the distilled reasoning. The resulting data is saved in Parquet format for efficient loading during training.

\subsection{RL Data}
The RL stage uses the \bench{} training and test splits. 
We identified a critical preprocessing issue for multiple-choice (MCQ) samples: 
in the original dataset, the ground-truth answer for MCQ problems is stored as a raw letter (e.g., \texttt{A}, \texttt{B}, \texttt{C}, or \texttt{D}) rather than in \verb|\boxed{}| format, causing the verifiable reward function to fail to match answers during RL training. To address this, we preprocess both the training and test sets by wrapping bare-letter MCQ ground-truth answers in \verb|\boxed{}|. 
Both datasets are stored in Parquet format for compatibility with the VeRL training framework.

\subsection{Model Configuration}
A subtle but critical issue arises when using \model{} directly for SFT. 
The Base model's default end-of-sequence token is \texttt{<|endoftext|>}, whereas the chat template used in SFT training data employs \texttt{<|im\_end|>} as the sequence delimiter. 
This mismatch causes the model to fail to recognize when a response should terminate, resulting in repetitive generation. 
To resolve this, we replace the Base model's \texttt{generation\_config.json} and \texttt{tokenizer\_config.json} with those from Qwen2.5-3B-Instruct, which properly defines \texttt{<|im\_end|>} as the end-of-sequence token. 
This configuration swap ensures that the model can correctly identify sequence boundaries during both SFT and subsequent RL training, without changing any model weights.

\section{Training Hyperparameters}
\Cref{tab:config} lists the hyperparameters used in our experiments. All RL variants share the same base configuration (group size, learning rate, batch size, training steps), and differ only in algorithm-specific parameters. For GRPO, we set the KL loss coefficient to 0.01 and the clipping ratio to 0.2. For GSPO, we use a tight clipping range with $\text{clip\_ratio\_low}=3\times10^{-4}$ and $\text{clip\_ratio\_high}=4\times10^{-4}$, and no KL penalty. For GMPO, we set the KL loss coefficient to 0.001 and the clipping ratio to 0.4.

\begin{table}[htbp]
  \centering
  \caption{Detailed training hyperparameters.}
  \label{tab:config}
  \begin{tabular}{lc}
    \toprule
    \multicolumn{2}{c}{\textbf{SFT}} \\
    \midrule
    Base model & \model{} \\
    SFT dataset & \sftdata{} (3,542 examples) \\
    Max response length & 768 \\
    Learning rate & 3e-6 \\
    Batch size & 36 \\
    Epochs & 1 \\
    Hardware & 4090*3 \\
    \midrule
    \multicolumn{2}{c}{\textbf{RL (shared)}} \\
    \midrule
    RL dataset & \bench{} train split \\
    Evaluation set & \bench{} test split \\
    RL algorithms & GRPO, GSPO, GMPO \\
    Group size & 4 \\
    Max response length & 768 \\
    Learning rate & 1e-6 \\
    Batch size & 18 \\
    Training steps & 1000 \\
    Hardware & 4090*3 \\
    \bottomrule
  \end{tabular}
\end{table}

\section{Extended Benchmark Comparison}
Table~\ref{tab:extended-results} provides an extended comparison with proprietary models, open-source general-purpose models, and open-source math-specialized models. We include the complete set of reported metrics to place the SignalReasoner experiments in a broader context.

\begin{table}[htbp]
  \centering
  \caption{Extended comparison on the \bench{} test set (\%).}
  \label{tab:extended-results}
  \begin{tabular}{lcccccc}
    \toprule
    Model & Size & MCQ & Fill-in & FEC & Overall & Output Tokens \\
    \midrule
    \multicolumn{7}{l}{\textit{Proprietary Models}} \\
    GPT-5 & -- & 63.91 & \textbf{63.20} & 41.36 & \textbf{57.87} & -- \\
    GPT-5-mini & -- & 67.67 & 53.99 & 40.31 & 53.00 & -- \\
    GPT-5-nano & -- & 57.14 & 37.82 & 30.37 & 39.25 & -- \\
    GPT-4o & -- & 54.14 & 43.62 & 24.61 & 40.37 & -- \\
    o4-mini & -- & 67.67 & 49.56 & 40.31 & 50.38 & -- \\
    Claude-4.0-Sonnet & -- & 60.15 & 56.30 & 42.93 & 53.75 & -- \\
    Gemini-2.5-Flash & -- & 63.16 & 56.09 & 43.46 & 54.25 & -- \\
    Gemini-2.5-Pro & -- & 66.17 & 50.42 & 36.65 & 49.75 & -- \\
    Grok-4-Fast & -- & \textbf{70.31} & 56.33 & 40.33 & 54.89 & -- \\
    \midrule
    \multicolumn{7}{l}{\textit{Open-Source General Models}} \\
    DeepSeek-R1 & 671B & 65.41 & 60.50 & 43.98 & 57.37 & -- \\
    DeepSeek-V3.1 & 671B & 66.17 & 58.85 & \textbf{45.03} & 56.87 & -- \\
    Llama-3.3-70B-Instruct & 70B & 54.14 & 38.03 & 28.27 & 38.37 & -- \\
    Qwen2.5-72B-Instruct & 72B & 51.88 & 35.50 & 32.46 & 37.50 & -- \\
    Qwen2.5-7B-Instruct & 7B & 39.10 & 21.85 & 26.18 & 25.75 & -- \\
    Gemma 3 27B & 27B & 42.11 & 30.04 & 27.75 & 31.50 & -- \\
    Gemma 3 12B & 12B & 36.84 & 21.43 & 21.99 & 24.12 & -- \\
    \midrule
    \multicolumn{7}{l}{\textit{Open-Source Math-Specialized Models}} \\
    Qwen2.5-Math-72B-Instruct & 72B & 60.15 & 40.55 & 33.51 & 42.13 & -- \\
    Qwen2.5-Math-7B-Instruct & 7B & 42.11 & 14.71 & 24.61 & 21.62 & -- \\
    DeepSeekMath-7B-RL & 7B & 43.61 & 13.66 & 25.65 & 21.50 & -- \\
    \midrule
    \multicolumn{7}{l}{\textit{SignalReasoner 3B Models}} \\
    Base & 3B & 28.57 & 9.03 & 9.42 & 12.37 & 907.41 \\
    Base-SFT & 3B & 42.11 & 26.68 & 24.61 & 28.75 & 604.45 \\
    A-GRPO & 3B & 57.89 & 30.88 & 27.23 & 34.50 & 668.33 \\
    A-GSPO & 3B & 61.65 & 30.88 & 26.18 & 34.88 & 279.25 \\
    A-GMPO & 3B & 51.88 & 32.98 & 26.70 & 34.63 & 395.84 \\
    B-GRPO & 3B & 49.62 & 33.61 & 25.65 & 34.38 & 600.10 \\
    B-GSPO & 3B & 57.89 & 35.08 & 30.37 & 37.75 & \textbf{243.59} \\
    B-GMPO & 3B & 54.14 & 38.03 & 31.41 & 39.12 & 258.87 \\
    \bottomrule
  \end{tabular}
\end{table}

The extended results reveal several patterns. First, the strongest proprietary and open-source general-purpose models achieve substantially higher overall accuracy than the 3B SignalReasoner variants. GPT-5 obtains the highest overall score among the listed external models at 57.87\%, while DeepSeek-R1 reaches 57.37\% and DeepSeek-V3.1 reaches 56.87\%. This gap is expected given the considerable differences in model scale and general mathematical capability, but it also highlights the difficulty of signal-domain mathematical reasoning for compact models.

Second, the SignalReasoner variants remain competitive with several smaller open-source baselines. B-GMPO achieves the highest overall accuracy among our 3B models at 39.12\%, exceeding Qwen2.5-7B-Instruct (25.75\%), Gemma~3 27B (31.50\%), and all three listed math-specialized baselines at 7B or 72B except Qwen2.5-Math-72B-Instruct (42.13\%). The category-level results also show that improvements are not limited to one question type: B-GMPO obtains the strongest SignalReasoner Fill-in score at 38.03\% and the strongest FEC score at 31.41\%, whereas A-GSPO obtains the highest MCQ score at 61.65\%.

Finally, the output-token statistics show a clear efficiency trade-off among the reinforcement-learning variants. B-GSPO produces the shortest average response, 243.59 tokens, while still reaching 37.75\% overall accuracy. B-GMPO is slightly longer at 258.87 tokens but achieves the best overall accuracy. In contrast, the GRPO variants generate substantially longer responses, with 668.33 and 600.10 tokens for A-GRPO and B-GRPO, respectively. These results support the observation that sequence-level optimization can improve response concision while preserving, and in the case of GMPO improving, answer accuracy.

\end{document}